\documentclass[a4paper,11pt]{article}
\usepackage[utf8]{inputenc}
\usepackage[T1]{fontenc}  
\usepackage{textcomp}          
\usepackage{amssymb,amsmath,amsfonts,epsfig}
\usepackage{algorithm}
\usepackage{algpseudocode}
\usepackage{amsthm}
\newtheoremstyle{myexample}
{12pt}
{12pt}
{\normalfont}
{}
{\bfseries}
{}
{1em}
{}
\theoremstyle{myexample}
\newtheorem{example}{Example}

\catcode`@=11 \@addtoreset{equation}{section}
\renewcommand\theequation{\thesection.\@arabic\c@equation}
\catcode`@=12

\def\theequation{\thesection.\arabic{equation}}
\usepackage{amsmath,amsfonts,amssymb}
\usepackage{booktabs}

\usepackage[title]{appendix}
\usepackage{bm}
\usepackage{algorithm}
\usepackage{amsmath,amsfonts,epsfig}
\usepackage{graphics}
\usepackage{supertabular}
\usepackage{amsthm}
\usepackage{mathtools}
\usepackage{amsfonts}
\usepackage{tikz}
\usepackage{pgfplots}
\usepackage{graphicx}
\usepackage{epstopdf}
\usepackage[numbers]{natbib}
\usepackage{float,epsfig, floatflt,here}
\usepackage{color}
\usepackage[colorlinks=true,citecolor=blue,linkcolor=blue]{hyperref}
\usepackage{fancyhdr}
\usepackage{etoolbox}
\usepackage{dsfont }
\usepackage{accents}

\usepackage{mathrsfs}
\usepackage{ upgreek }

\newcommand{\tnorm}[1]{{\left\vert\kern-0.25ex\left\vert\kern-0.25ex\left\vert #1\right\vert\kern-0.25ex\right\vert\kern-0.25ex\right\vert}}

\newcommand{\vertiii}[1]{{\left\vert\kern-0.25ex\left\vert\kern-0.25ex\left\vert #1
		\right\vert\kern-0.25ex\right\vert\kern-0.25ex\right\vert}}

\renewcommand{\theequation}{A.\arabic{equation}}
\numberwithin{equation}{section}
\numberwithin{figure}{section}
\numberwithin{table}{section}

\begin{document}
	\title{PINNs Algorithmic Framework for Simulation of Nonlinear Burgers' Type Models}
	\author{  Ajeet Singh\thanks{ Department of Mathematics, Indian Institute of Technology, Roorkee, Roorkee-247667, India (ajeet\_s@ma.iitr.ac.in)} \and  Ram Jiwari \thanks{Department of Mathematics, Indian Institute of Technology, Roorkee, Roorkee-247667, India (ram.jiwari@ma.iitr.ac.in)}  \and Vikram\thanks{Department of Mathematics, Indian Institute of Technology, Roorkee, Roorkee-247667, India (vikramaryasaharan@gmail.com)} \and Ujjwal Saini
		\thanks {Department of Computer Science and Engineering, Graphic Era University (deemed to be) Dehradun, India
			(sainiujjwalus@gmail.com)}}
	\date{}
	\maketitle
	\begin{abstract}
		In this work, a physics-informed neural networks (PINNs) based algorithm is used for simulation of nonlinear 1D and 2D Burgers' type models. This scheme relies on a neural network built to approximate the problem solution and use a trial function that meets the initial data and boundary criteria. First of all, a brief mathematical formulation of the problem and the structure of PINNs, including the neural network architecture, loss construction, and training methodology is described. Finally, the algorithm is demonstrated with five test problems involving variations of the 1D coupled, 2D single and 2D coupled Burgers' models. We compare the PINN-based solutions with exact results to assess accuracy and convergence of the developed algorithm. The results demonstrate that PINNs may faithfully replicate nonlinear PDE solutions and offer competitive performance in terms of inaccuracy and flexibility. This work demonstrates the potential of PINNs as a reliable approach to solving complex time-dependent PDEs.
		
	\end{abstract}
	
	\section{Introduction}
	PDEs, are the cornerstone of mathematical modeling and applied mathematics.  Over the years, nonlinear PDEs in fluid mechanics and heat transfer have been the focus of much research, and as more practical engineering considerations are made, their complexity has grown \cite{moseley2023finite,ding2024artificial}. Among these, Burgers' equation has emerged as a fundamental model combining nonlinear propagation and diffusive effects, serving as an important benchmark for numerical methods \cite{burgers2013nonlinear,jiwari2022local}. Its broad applications across gas dynamics, heat conduction \cite{mittal2012differential,jiwari2019meshfree}, elasticity \cite{jiwari2015hybrid}, and groundwater solute transport \cite{jiwari2023analysis,jiwari2012haar} have motivated extensive analytical and numerical studies.
	
	Traditional numerical methods for Burgers' equation solutions include the FEM \cite{kumar2025error,singh2025high,jiwari2024finite}, FDM \cite{iserles2009first}, and FVM \cite{sheng2018finite}. Among these, FEM has emerged as the gold standard for many applications due to its capacity to manage intricate geometries through variational formulations and unstructured meshing \cite{bungartz2004sparse}. However, FEM and related mesh-based techniques face significant challenges in high-dimensional problems due to the curse of dimensionality [12], and may struggle with irregular domains or solutions containing singularities \cite{egger2014energy}. The method's implementation, while supported by robust computational frameworks \cite{alnaes2011fenics,blatt2016distributed}, remains non-trivial for many practical applications.
	
	Recently, PINNs have become as a promising alternative algorithm \cite{raissi2019physics} for simulating the PDEs. By leveraging deep learning architectures and automatic differentiation \cite{baydin2018automatic}, PINNs offer a mesh-free methodology that potentially circumvents the dimensionality limitations of classical techniques \cite{wojtowytsch2020can}. The fundamental concept involves training neural networks to minimize the residual of the governing PDE along with its boundary and initial data \cite{raissi2019physics}. This method has been expanded through several specialized formulations \cite{kharazmi2019variational,jagtap2020conservative} and has shown success across a variety of PDE classes \cite{smith2022hyposvi}.
	Because of their versatility and capacity to directly integrate physical laws into the learning process,  PINNs have found use in a broad variety of scientific and engineering domains. PINNs have been applied to both forward and inverse issues in fluid dynamics, providing a mesh-free option for intricate flow simulations~\cite{mao2020physics}. They make it easier to model electromagnetic wave propagation and field distributions in electromagnetism~\cite{kovacs2022conditional}. In order to represent the stress-strain behavior of complicated structures, PINNs are also useful for modeling the deformation of elastic materials~\cite{li2021physics}. They aid in the characterisation of subsurface structures and the modeling of seismic wave propagation in seismology~\cite{smith2022hyposvi}.
	
	Due to their adaptability, PINNs can be utilized in a variety of physical systems with only minor modifications to the network architecture. Their universality is one of the primary advantages that has prompted their adoption in interdisciplinary research. Their suitability for complex domains, non-smooth PDEs, and problems requiring efficient training or parallelization is further enhanced by recent advancements such as variational PINNs (VPINNs)~\cite{kharazmi2019variational}, extended PINNs (XPINNs)~\cite{jagtap2020extended}, and conservative PINNs (cPINNs)~\cite{jagtap2020conservative}. The development of user-friendly software tools like DeepXDE~\cite{lu2021deepxde}, NVIDIA Modulus~\cite{hennigh2021nvidia}, and NeuroDiffEq~\cite{chen2020neurodiffeq} has significantly lowered the entry barrier for implementing PINNs, making them accessible for researchers and practitioners across various domains.
	The key novelties of this study are summarized below:
	
	(i) This work extends the application of PINNs to both 1D and 2D Burgers’ models, demonstrating the model’s flexibility and generalization capabilities across different spatial dimensions and coupled nonlinear dynamics.
	
	(ii) By incorporating the governing PDEs directly into the loss function, the method achieves accurate solutions without relying on mesh-based spatial discretization, offering a fully differentiable and mesh-free alternative to classical numerical solvers.
	
	(iii) The accuracy and stability of the PINN approach are rigorously validated using analytical solutions and quantified through \( L^\infty \) and \( L^2 \) norm evaluations, reinforcing its potential as a reliable solver for nonlinear PDEs under data-scarce conditions.
	
	This work is designed as: In Section~\ref{sec:background}, we present the mathematical background, including the governing PDEs and relevant analytical concepts. Section~\ref{sec:pinn_intro} introduces Physics-Informed Neural Networks (PINNs), highlighting their motivation and core principles. The key components of the PINN framework, such as the loss formulation and network architecture, are detailed in Section~\ref{sec:components}. In Section~\ref{sec:algorithm}, we describe the algorithmic implementation and apply the PINN method to three different example of coupled Burgers' model. 
	For each case, we compare the PINN-based solution with the exact or reference solution to assess accuracy and efficiency. Finally, in Section~\ref{sec:conclusion}, we offer a few closing thoughts about the obtained results and give some directions for future study.
	
	\section{Mathematical Background of the Model} \label{sec:background}
	Herein, we study a particular type of model the PDEs  that takes the general form
	\begin{equation}
		A(u(x,t)) = f(x,t), \quad x \in \Omega \subset \mathbb{R}^d, \quad t \in [0,T],
	\end{equation}
	where 
	\[
	u : \Omega \times [0,T] \rightarrow \mathbb{R}^n
	\]
	denotes the vector-valued solution of the system, $A$ is a (possibly nonlinear) differential operator, and
	\[
	f : \Omega \times [0,T] \rightarrow \mathbb{R}^n
	\]
	is a given source term. To ensure a well-posed problem, we complement the system with suitable boundary data (BD)and initial data (ID):
	\begin{equation}
		B(u(x,t)) = g(x,t), \quad u(x,0) = h(x), \quad x \in \Omega.
	\end{equation}
	
	In this work, we will simulate the nonlinear Burgers’ types system with convective and diffusive characteristics with the help of PINNs based algorithm. The general form of the  model is represented as: \cite{hussein2020weak}:
	\begin{align}
		u_t + u \cdot \nabla u - \nu \Delta u &= f_1(x,t), \\
		v_t + v \cdot \nabla v - \nu \Delta v &= f_2(x,t),
		\quad x \in \Omega \subset \mathbb{R}^d, \quad t \in [0,T],
	\end{align}
	where $u(x,t), v(x,t)$ are scalar or vector fields (depending on the context), $\nu > 0$ is the viscosity coefficient, and $f_1, f_2$ are source terms.
	The nonlinearity in the convective terms $u \cdot \nabla u$ and $v \cdot \nabla v$ characterizes the transport behavior, while the Laplacian $\Delta$ accounts for diffusive effects.
	To be well-defined, the system requires ID:
	\begin{equation}
		u(x,0) = u_0(x), \quad v(x,0) = v_0(x),
	\end{equation}
	and some appropriate BCs depending on the physical setting.
	
	The Burgers’ system serves as a prototype model in fluid dynamics and nonlinear wave propagation, and is often used to test numerical methods due to its rich structure combining nonlinearity and diffusion. It generalizes the classical scalar Burgers’ equation (see \cite{mittal2014collocation}) and appears in various applications including traffic flow, acoustics, and turbulence modeling.
	\section{Brief Overview of PINNs} \label{sec:pinn_intro}
	
	
	PINNs are designed to approximate solutions of  PDEs by embedding the underlying physical laws into the structure of a neural network. The solution is seen as the output of a neural network, where the input consists of the independent variables, such as spatial and temporal coordinates. For the coupled Burgers' equation, the PINN approach involves constructing a neural network that approximates the solution components \( u(x,t) \) and \( v(x,t) \) simultaneously.
	
	Instead of relying on a discretized mesh, PINNs utilize randomly sampled points—known as collocation points—from the space-time domain. These points are used to enforce the governing equations, initial data (ID) and boundary data (BD) through a unified loss function. The total loss typically includes three main components: the PDE residuals, the ID errors, and the BD errors. These residuals are produced via automatic differentiation, allowing for precise and efficient assessment of the derivatives involved in the governing system. With \( N_l \) hidden layers and \( N_e^{(l)} \) neurons per layer \( l \), a fully connected feed-forward neural network is employed. The architecture is flexible and can be altered to accommodate the complexity of the linked system. The network is trained by minimizing the composite loss function using gradient-based optimization methods. It is common practice to employ a hybrid training strategy, beginning with the Adam optimizer for coarse convergence and progressing to the L-BFGS optimizer for high-precision outcomes.
	
	To enhance generalization and ensure a thorough exploration of the solution space, Latin Hypercube Sampling (LHS) is often used to generate collocation points. These points may be re-sampled at each epoch to improve domain coverage. The PINN framework thus provides a continuous representation of the solution that satisfies both the physics of the coupled Burgers' system and any available data, without requiring an explicit mesh or discretization of the domain.
	
	\subsection{Design and Depth of the PINNs }
	
	The PINN created for fully connected feedforward neural network that represents the solution to the coupled Burgers' system by mapping temporal and spatial inputs \( (x, t) \) to two outputs \( (u, v) \). The architecture is modular, and a hyperparameter tuning phase is used to choose the width (number of neurons per layer) and depth (number of layers).\\
	The network consists of the following components \cite{zhai2023deep}:
	\begin{itemize}
		\item \textbf{Input Layer:} Takes two scalar inputs: the spatial coordinate \( x \in [0, 1] \) and temporal coordinate \( t \in [0, 1] \).
		\item \textbf{Hidden Layers:} A sequence of \( L \in \{3, 4, 5, 6, 7\} \) fully connected layers, each comprising \( H \in \{20, 30, 40, 50, 60\} \) neurons. Each hidden layer applies a linear transformation followed by a Tanh activation function:
		\[
		z^{(l+1)} = \tanh(W^{(l)} z^{(l)} + b^{(l)})
		\]
		\item \textbf{Output Layer:} A final linear transformation maps the hidden features to two scalar outputs:
		\[
		\begin{bmatrix}
			u(x, t) \\
			v(x, t)
		\end{bmatrix}
		\]
	\end{itemize}
	The full network is trained end-to-end with the help of the Adam optimizer. During training, gradients of the output with respect to \( x \) and \( t \) are computed using automatic differentiation to evaluate the residuals of the PDEs.
	
	The optimal architecture, i.e., the best performing combination of \( L \) and \( H \), was chosen through a systematic hyperparameter sweep based on minimizing both \( L^2 \) and \( L^\infty \) error norms on validation sets.
	
	\begin{figure}[h]
		\centering
		\includegraphics[width=0.8\textwidth]{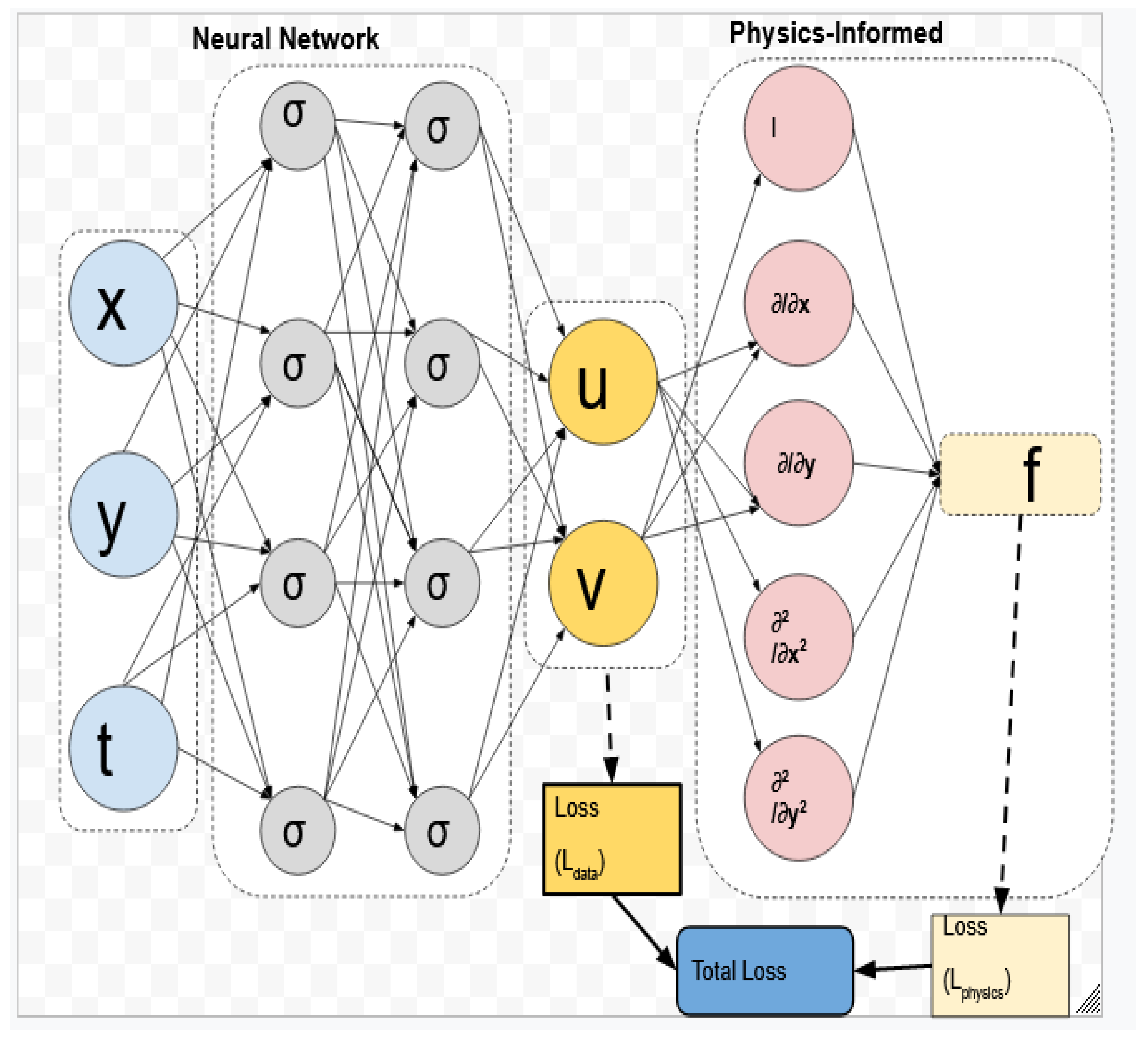}
		\caption{Schematic of the PINN architecture: Input \((x,t)\) passes through several fully connected layers with Tanh activations, and outputs the approximated functions \( u(x,t) \) and \( v(x,t) \).}
		\label{fig:pinn_architecture}
	\end{figure}
	\section{ PINNs Model and their Components} \label{sec:components}
	The training objective integrates several mathematical and algorithmic components essential for learning a solution that satisfies the general coupled Burgers' system \cite{karniadakis2021physics,thanasutives2021adversarial}:
	\[
	\begin{aligned}
		\frac{\partial u}{\partial t} + \alpha u \frac{\partial u}{\partial x} + \beta v \frac{\partial u}{\partial x} &= \epsilon \frac{\partial^2 u}{\partial x^2}, \\
		\frac{\partial v}{\partial t} + \gamma v \frac{\partial v}{\partial x} + \delta u \frac{\partial v}{\partial x} &= \epsilon \frac{\partial^2 v}{\partial x^2}
	\end{aligned}
	\]
	
	\subsection*{Activation Function:} The choice of activation function in PINNs plays a crucial role in determining the model’s ability to approximate nonlinear PDE solutions accurately. In the discussed framework, the hyperbolic tangent (tanh) activation function is employed across all hidden layers. Below, we elaborate on its significance, alternatives, and impact on training dynamics. 
	The  \( \tanh \) function, defined as:
	\[
	\tanh(x) = \frac{e^x - e^{-x}}{e^x + e^{-x}}
	\]
	This activation introduces nonlinearity, enabling the neural network to approximate complex, smooth functions needed to represent solutions to nonlinear PDEs.
	
	\subsection*{Physics-Informed Loss Function:} The loss function is constructed to enforce both data fidelity and physical consistency. It consists of three primary components.
	\begin{itemize}
		\item[(a)]\textbf{PDE Residual Loss:} In the specific problem considered:
		\[
		f_u = u_t - \epsilon u_{xx} + 2uu_x - (uv)_x, \quad f_v = v_t - \epsilon v_{xx} + 2vv_x - (uv)_x
		\]
		The residuals are evaluated at collocation points and minimized using the mean squared error:
		\[
		\mathcal{L}_{\text{PDE}} = \frac{1}{N_r} \sum_{i=1}^{N_r} \left( f_u(x_i,t_i)^2 + f_v(x_i,t_i)^2 \right)
		\]
		\item[(b)] \textbf{Initial Condition Loss:} The network is penalized for deviating from the known initial profile:
		\[
		u(x,0) = v(x,0) = \cos(\pi x)
		\]
		The loss is computed as:
		\[
		\mathcal{L}_{\text{IC}} = \frac{1}{N_0} \sum_{i=1}^{N_0} \left( \| u(x_i,0) - u_0(x_i) \|^2 + \| v(x_i,0) - v_0(x_i) \|^2 \right)
		\]
		\item 
		[(c)]\textbf{Boundary Condition Loss:} For BCs at \( x=0 \) and \( x=1 \), a similar mean squared error term:
		\[
		\mathcal{L}_{\text{BC}} = \frac{1}{N_b} \sum_{i=1}^{N_b} \left( \| u(x_i,t_i) - u_{\text{BC}}(x_i,t_i) \|^2 + \| v(x_i,t_i) - v_{\text{BC}}(x_i,t_i) \|^2 \right)
		\]
	\end{itemize}
	The training process optimizes a composite loss function that balances adherence to physical laws with the given ICs and BCs. The total loss is formulated as:
	\begin{equation}
		\mathcal{L}{\text{total}} = \mathcal{L}{\text{PDE}} + \lambda_{\text{IC}} \mathcal{L}{\text{IC}} + \lambda{\text{BC}} \mathcal{L}_{\text{BC}}
	\end{equation}
	The composite loss function combines three key components: the partial differential equation residual $\mathcal{L}{\text{PDE}}$, the initial condition term $\mathcal{L}{\text{IC}}$, and the boundary condition term $\mathcal{L}{\text{BC}}$. Weighting coefficients $\lambda{\text{IC}}$ and $\lambda_{\text{BC}}$, which are typically set to 10, balance the relative importance of each component in the optimization process. This weighting strategy ensures that the governing physics and the problem-specific constraints are appropriately prioritized by preventing any one component from driving the training process. The careful selection of these weights proves particularly important when working with limited training data for initial and boundary conditions.
	\subsection*{Optimization Method}
	The model training utilizes the Adam optimizer~\cite{kingma2014adam}, which computes 
	adaptive learning rates for each parameter. The update rule is:
	
	\begin{equation}
		\theta_{t+1} = \theta_t - \eta \frac{\hat{m}_t}{\sqrt{\hat{v}_t} + \epsilon}
	\end{equation}
	
	where $\hat{m}_t = m_t/(1-\beta_1^t)$ and $\hat{v}_t = v_t/(1-\beta_2^t)$ are bias-corrected 
	moment estimates, with $m_t$ and $v_t$ being exponential moving averages of the gradient 
	and squared gradient respectively. Default values $\beta_1=0.9$, $\beta_2=0.999$, 
	$\eta=10^{-3}$, and $\epsilon=10^{-8}$ were used throughout our experiments.
	Although no explicit \( L^2 \) regularization term is added, early stopping and weighted loss terms serve as implicit regularizers.
	
	\noindent \textbf{Hyperparameter Tuning:} The number of layers \( L \) and neurons per layer \( H \) are determined via a grid search to minimize both \( L^2 \) and \( L^\infty \) validation errors:
	\[
	\| u_{\text{pred}} - u_{\text{exact}} \|_2, \quad \| u_{\text{pred}} - u_{\text{exact}} \|_\infty
	\]
	These components collectively ensure that the trained neural network approximates the true solution of the generalized coupled Burgers' system in a data-efficient and physics-consistent manner.
	\section{PINN Algorithm and Numerical Results} \label{sec:algorithm}
	\noindent\fbox{\parbox{\textwidth}{
			\begin{enumerate}
				\item \textbf{Initialize the Network:} Define a feedforward neural network with \( L \) hidden layers and \( H \) neurons per layer.
				
				\item \textbf{Sample Training Points:}
				\begin{itemize}
					\item Collocation points \( (x_r, t_r) \) in the interior of the domain
					\item Initial condition points \( (x_0, t=0) \)
					\item Boundary condition points \( (x_b, t_b) \) at spatial domain edges
				\end{itemize}
				
				\item \textbf{Define Loss Functions:}
				\begin{itemize}
					\item PDE residual loss using automatic differentiation
					\item Mean squared error for initial and boundary condition enforcement
				\end{itemize}
				
				\item \textbf{Train the Network:} For \( E \) epochs, repeat:
				\begin{itemize}
					\item Perform forward pass to evaluate \( u(x,t), v(x,t) \)
					\item Compute total loss:
					\[
					\mathcal{L}_{\text{total}} = \mathcal{L}_{\text{PDE}} + 10\mathcal{L}_{\text{IC}} + 10\mathcal{L}_{\text{BC}}
					\]
					\item Backpropagate gradients and update weights using Adam optimizer
				\end{itemize}
				
				\item \textbf{Evaluate the Trained Model:}
				\begin{itemize}
					\item Predict \( u, v \) at test points
					\item Compare with exact solution using \( L^2 \) and \( L^\infty \) norms
					\item Visualize predictions and error metrics
				\end{itemize}
			\end{enumerate}
	}}\\\\
	\begin{example}[ \textbf{Coupled Burgers' Equations}]:\label{ex:prob1} The coupled system is simulated over the domain \( x \in (-\pi, \pi) \) with analytic solution \( u(x,t) = v(x,t) = e^{-t}\sin(x) \)\cite{mittal2012differential}. The considered coupled model is as
		\begin{align}
			u_t - u_{xx} - 2u u_x + (uv)_x &= 0,\nonumber \\
			v_t - v_{xx} - 2v v_x + (uv)_x &= 0.\label{1}
		\end{align}
		The results of the models are reported in Table 5.1 and Figs 5.1-5.2. 
		\begin{table}[h!]
			\centering
			\begin{tabular}{|c|cc|cc|}
				\hline
				\textbf{t} & \multicolumn{2}{c|}{\textbf{error\_u}} & \multicolumn{2}{c|}{\textbf{error\_v}} \\ \hline
				& \( L^\infty \) & \( L^2 \) & \( L^\infty \) & \( L^2 \) \\ \hline
				0.5  & 7.4050e-03 & 3.3142e-03 & 8.7591e-03 & 3.8617e-03 \\ \hline
				1.0  & 1.0338e-02 & 4.3086e-03 & 1.0449e-02 & 4.2649e-03 \\ \hline
				5.0  & 6.4977e-03 & 3.8595e-03 & 5.3947e-03 & 2.9185e-03 \\ \hline
				10.0 & 3.4999e-03 & 1.8519e-03 & 3.5661e-03 & 1.8522e-03 \\ \hline
			\end{tabular}
			\caption{Error metrics for Example 1.}\label{tab:prob1_errors}
		\end{table}
		\begin{figure}[h!]
			\centering
			\includegraphics[width=0.95\textwidth]{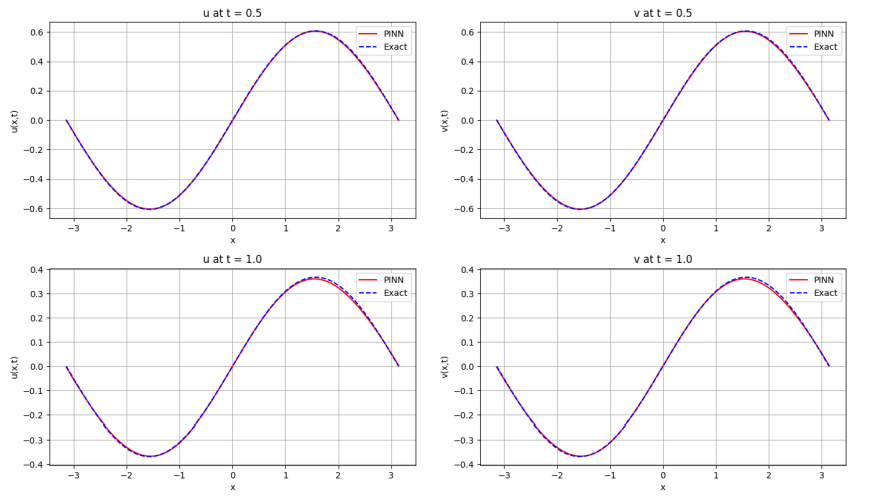}
			\caption{Comparison at \( t=0.5 \) and \( t=1.0 \) for Example 1.}\label{fig:prob1_compare}
		\end{figure}
		The errors peak at \( t=1.0 \) then gradually decrease, with \( v \) showing slightly higher errors than \( u \) throughout the simulation. The \( L^\infty \) errors dominate the \( L^2 \) norms, indicating localized discrepancies. The predicted solutions (dashed lines) are very analogous with exact solutions (solid lines) at both time levels. Minor deviations appear near the boundaries \( x=\pm\pi \), consistent with the error metrics in Table~\ref{tab:prob1_errors}. The maximum error occurs at \( t=1.0 \), matching the tabulated data.
		
		\begin{figure}[h!]
			\centering
			\includegraphics[width=0.95\textwidth]{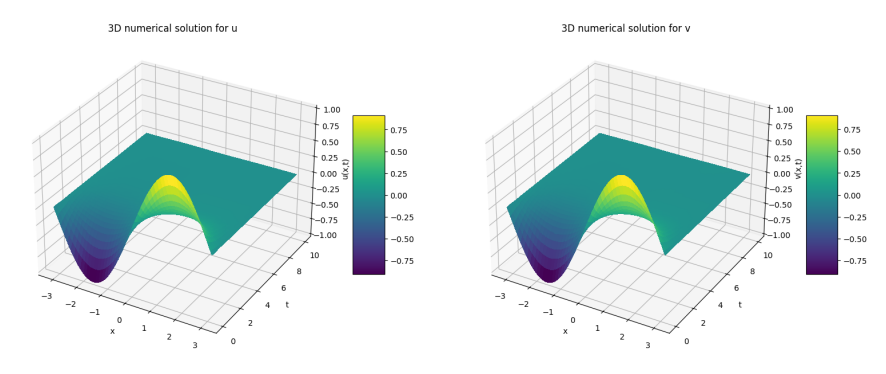}
			\caption{3D space-time solution for Test Problem 1}\label{fig:prob1_3d}
		\end{figure}
		The 3D visualization shown in Fig 5.2 confirms the expected exponential decay \( e^{-t} \) and preserves the sinusoidal spatial pattern accurately. No numerical instabilities are observed across the space-time domain, demonstrating the method's stability for coupled nonlinear systems.
	\end{example}
	\begin{example}[\textbf{ Modified Burgers' Equations:}]\label{ex:prob3}
		We consider the equation as follows \cite{bak2019semi}:
		\begin{align*}
			u_t - u_{xx} - 2u u_x + 0.1(uv)_x &= 0, \\
			v_t - v_{xx} - 2v v_x + 0.3(uv)_x &= 0, \quad x \in (-10, 10), \; t > 0,
		\end{align*}
		where the boundary conditions are extracted from the exact solution.
		\begin{align*}
			u(x,t) &= 0.05 \left(1 - \tanh\left(-0.00625 (x + 0.0125 t)\right)\right), \\
			v(x,t) &= 0.05 \left(-0.5 - \tanh\left(-0.00625 (x + 0.0125 t)\right)\right).
		\end{align*}
		Errors shown in Table 5.2 of the problem are an order of magnitude smaller than in problems 1--2, with \( v \) consistently showing higher errors than \( u \). The \( L^2 \) errors decrease over time, indicating good long-term stability.
		The tanh-wave propagates correctly without dispersion artifacts. Boundary effects are minimal despite the large domain, and the solution maintains its shape throughout the simulation. 
		The simulated solutions by PINNs algorithm of the model are plotted very nicely in Fig. 5.3. 
		
		\begin{table}[h]
			\centering
			\caption{Comparison of $L^\infty$ and $L^2$ errors for $u$ across different methods}
			\label{tab:error_comparison_u}
			\begin{tabular}{c|cc|cc|c|c|cc}
				\hline
				$t$ & \multicolumn{2}{c|}{\textbf{Bak et al. \cite{bak2019semi}}} & \multicolumn{2}{c|}{\textbf{Ahmad et al. \cite{ahmad2019numerical}}}&\multicolumn{1}{c|}{\textbf{Khater \cite{khater2008chebyshev}}}&\multicolumn{1}{c|}{\textbf{Rashid \cite{rashid2009fourier}}} & \multicolumn{2}{c}{\textbf{PINN (error\_u)}} \\
				& $L^\infty$ & $L^2$ & $L^\infty$ & $L^2$ & $L^\infty$ &$L^\infty$ &$L^\infty$ & $L^2$ \\
				\hline
				0.5  & 5.6194e-4 & 5.7744e-4 & 2.1840e-4 & 2.2440e-4& 1.44e-3&9.619e-4& 3.2962e-4 & 2.5975e-4 \\
				1.0  & 1.2083e-3 & 2.3460e-4 & 4.3681e-4 & 4.4886e-4 & 1.27e-3& 1.153e-3& 3.2357e-4 & 2.6682e-4 \\
				5.0  & 6.0644e-3 & 6.0584e-3 & 2.1844e-3 & 2.2464e-3 &-&-& 4.1237e-4 & 2.4964e-4 \\
				10.0 & 1.1633e-2 & 1.1365e-2 & 4.3694e-3 & 4.4981e-3 &-&-& 2.5434e-4 & 1.5171e-4 \\
				\hline
			\end{tabular}
		\end{table}
		
		\begin{table}[h]
			\centering
			\caption{Comparison of $L^\infty$ and $L^2$ errors for $v$ across different methods}
			\label{tab:error_comparison_v}
			\begin{tabular}{c|cc|cc|c|c|cc}
				\hline
				$t$ & \multicolumn{2}{c|}{\textbf{Bak et al. \cite{bak2019semi}}} & \multicolumn{2}{c|}{\textbf{Ahmad et al. \cite{ahmad2019numerical}}}&\multicolumn{1}{c|}{\textbf{Khater \cite{khater2008chebyshev}}}&\multicolumn{1}{c|}{\textbf{Rashid \cite{rashid2009fourier}}} & \multicolumn{2}{c}{\textbf{PINN (error\_u)}} \\
				& $L^\infty$ & $L^2$ & $L^\infty$ & $L^2$ & $L^\infty$ &$L^\infty$ &$L^\infty$ & $L^2$ \\
				\hline
				0.5  & 4.8567e-4 & 4.3123e-4 & 2.5169e-4 & 2.5955e-4 & 5.42e-4& 3.332e-4& 6.4322e-4 & 6.0488e-4 \\
				1.0  & 1.0169e-3 & 9.1399e-4 & 5.0343e-4 & 5.1920e-4 & 1.29e-3& 1.162e-3&  6.5330e-4 & 6.0512e-4 \\
				5.0  & 4.6046e-3 & 4.3400e-3 & 2.5190e-3 & 2.5996e-3  &-&-&  7.0377e-4 & 5.2899e-4 \\
				10.0 & 8.2233e-3 & 7.9933e-3 & 5.0426e-3 & 5.2084e-3 &-&-& 6.3846e-4 & 4.9401e-4\\
				\hline
			\end{tabular}
		\end{table}
		The PINN method demonstrates significantly lower $L^\infty$ and $L^2$ errors for both $u$ and $v$ compared to existing methods. This consistent accuracy across all time levels highlights the efficiency and robustness of the PINN approach.
		In brief, PINN offers a more effective results than \cite{khater2008chebyshev,rashid2009fourier,bak2019semi,ahmad2019numerical}.

		\begin{figure}[h!]
			\centering
			\includegraphics[width=0.95\textwidth]{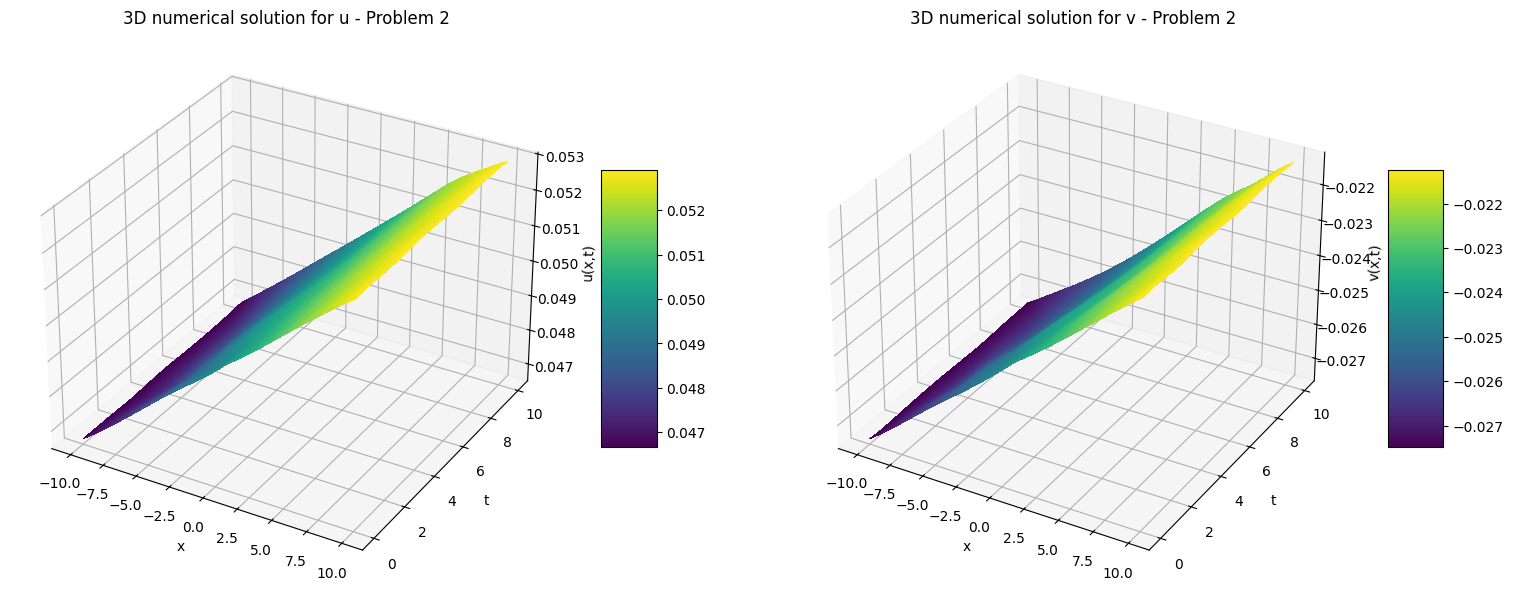}
			\caption{3D space-time solution for Example 2.}\label{fig:prob3_3d}
		\end{figure}
		
	\end{example}
	\begin{example}[\textbf{ High-Reynolds-Number Regime}]:\label{ex:prob2}
		For \( x \in (0,1) \) with \( \text{Re}=10^6 \) and exact solution \( u(x,t) =v(x,t)= e^{-\epsilon \pi^2 t}\cos(\pi x) \)\cite{wazwaz2007multiple}, the following model:
		\begin{align*}
			u_t - \epsilon u_{xx} + 2u u_x - (uv)_x &= 0, \\
			v_t - \epsilon v_{xx} + 2v v_x - (uv)_x &= 0.
		\end{align*}
		is simulated by the proposed algorithm. Here, $\epsilon=\frac{1}{R}$ is considered. The computed results were shown in Fig. \ref{fig:prob2_early}-\ref{fig:prob2_3d}.
		
		\begin{figure}[h!]
			\centering
			\includegraphics[width=0.95\textwidth]{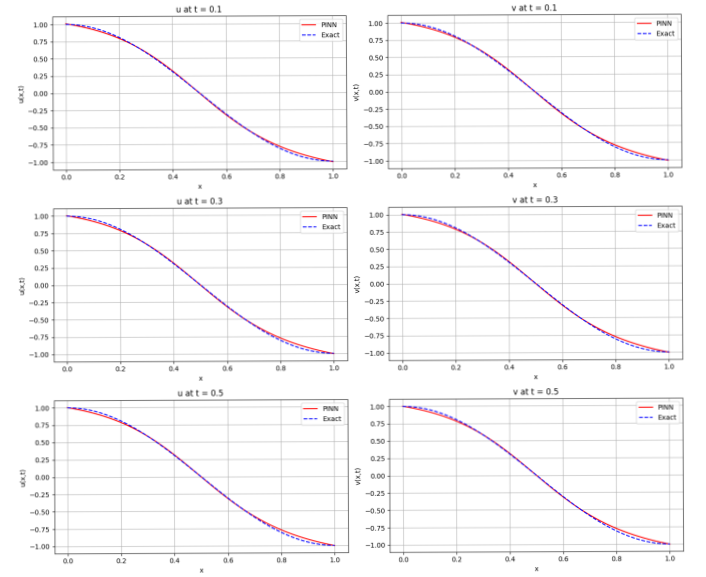}
			\includegraphics[width=0.95\textwidth]{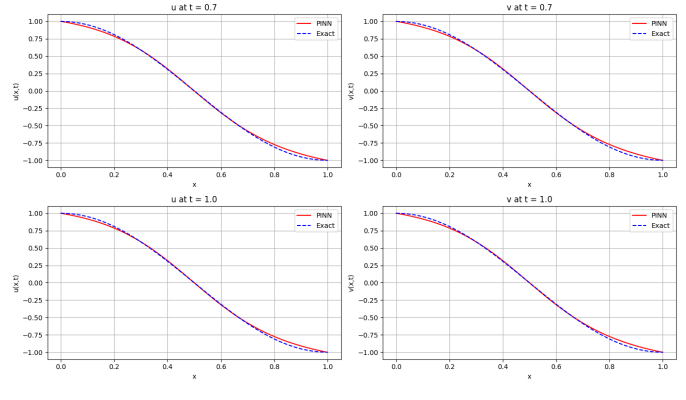}
			\caption{Solutions at \( t=0.1, 0.3, 0.5, 0.7, 1.0 \) of Example 3.}
			\label{fig:prob2_early}
		\end{figure}
		
		The predictions maintain good agreement with the exact solutions despite the high Reynolds number. Small phase shifts become visible at \( t=0.5 \), particularly near \( x = 0.5 \). Maximum deviations occur at \( t=0.7 \), where the solution gradients are sharpest. By \( t=1.0 \), the predictions stabilize while maintaining the correct cosine profile.
		
		\begin{figure}[h!]
			\centering
			\includegraphics[width=0.95\textwidth]{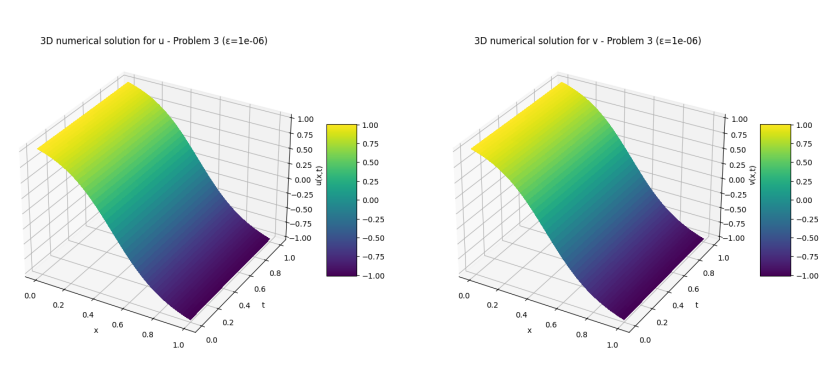}
			\caption{3D space-time solution of Example 3}
			\label{fig:prob2_3d}
		\end{figure}
		
		The 3D plot in Fig.~\ref{fig:prob2_3d} confirms proper exponential decay and resolves sharp gradients without numerical oscillations. The method demonstrates robustness in high-Reynolds-number regimes where traditional schemes often fail.
		
	\end{example}
	\begin{example}(\textbf{2D Burgers' model:})
		Herein, consider single non-linear 2D Burgers' model as $$u_t(x, y, t) = \frac{1}{R} \left[ u_{xx}(x, y, t) + u_{yy}(x, y, t) \right] - u\,u_x(x, y, t) - u\,u_y(x, y, t)
		$$
		
		with ID and BD taken from the given solutions over the domain $[0,1]\times[0,1]$ :\[u(x,y,t)=\frac{1}{1+e^{\frac{R(x+y-t)}{2}}}\]
		\begin{figure}[h!]
			\centering
			\begin{minipage}{0.48\textwidth}
				\centering
				\includegraphics[width=\textwidth]{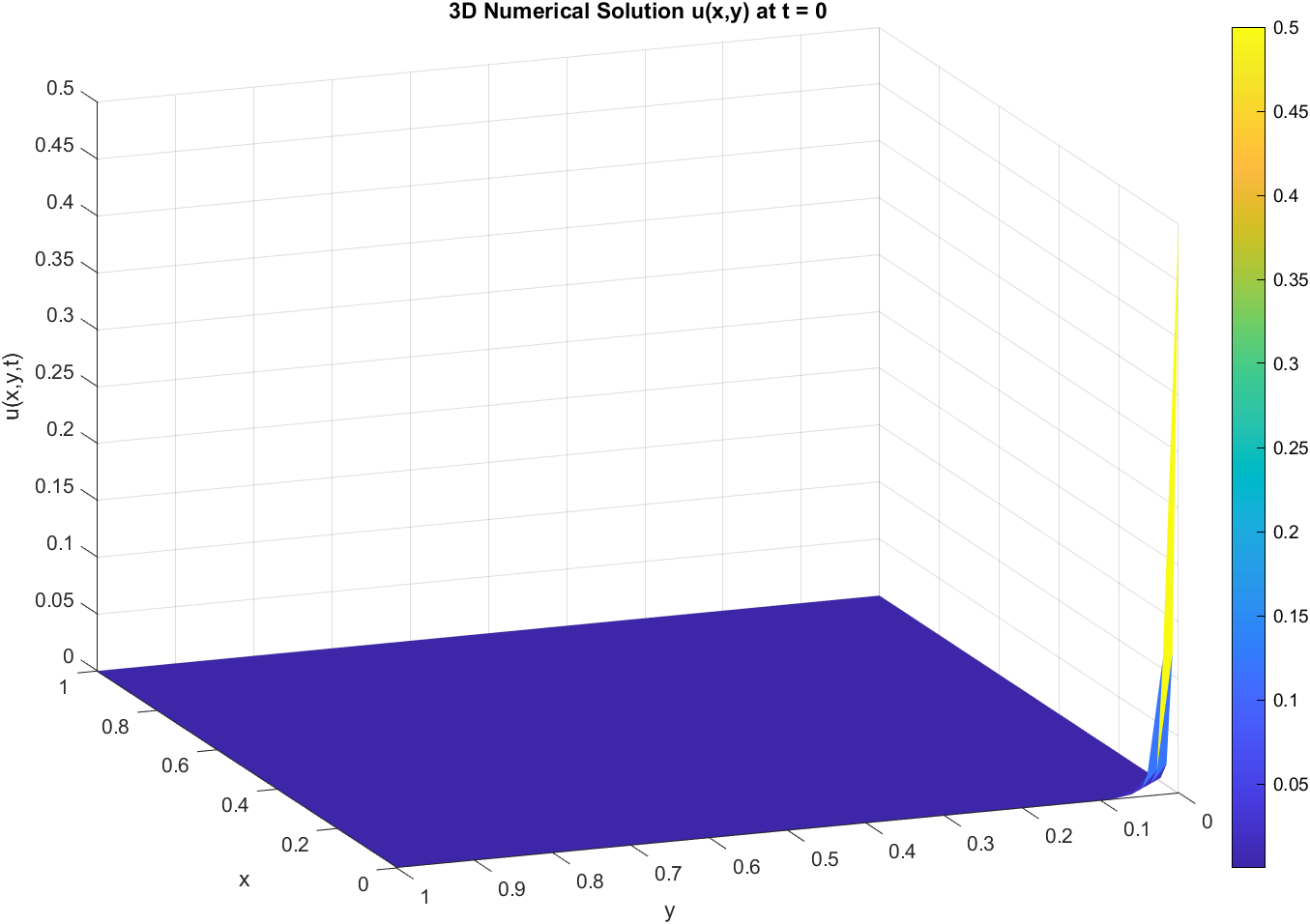}
			\end{minipage}
			\hfill
			\begin{minipage}{0.48\textwidth}
				\centering
				\includegraphics[width=\textwidth]{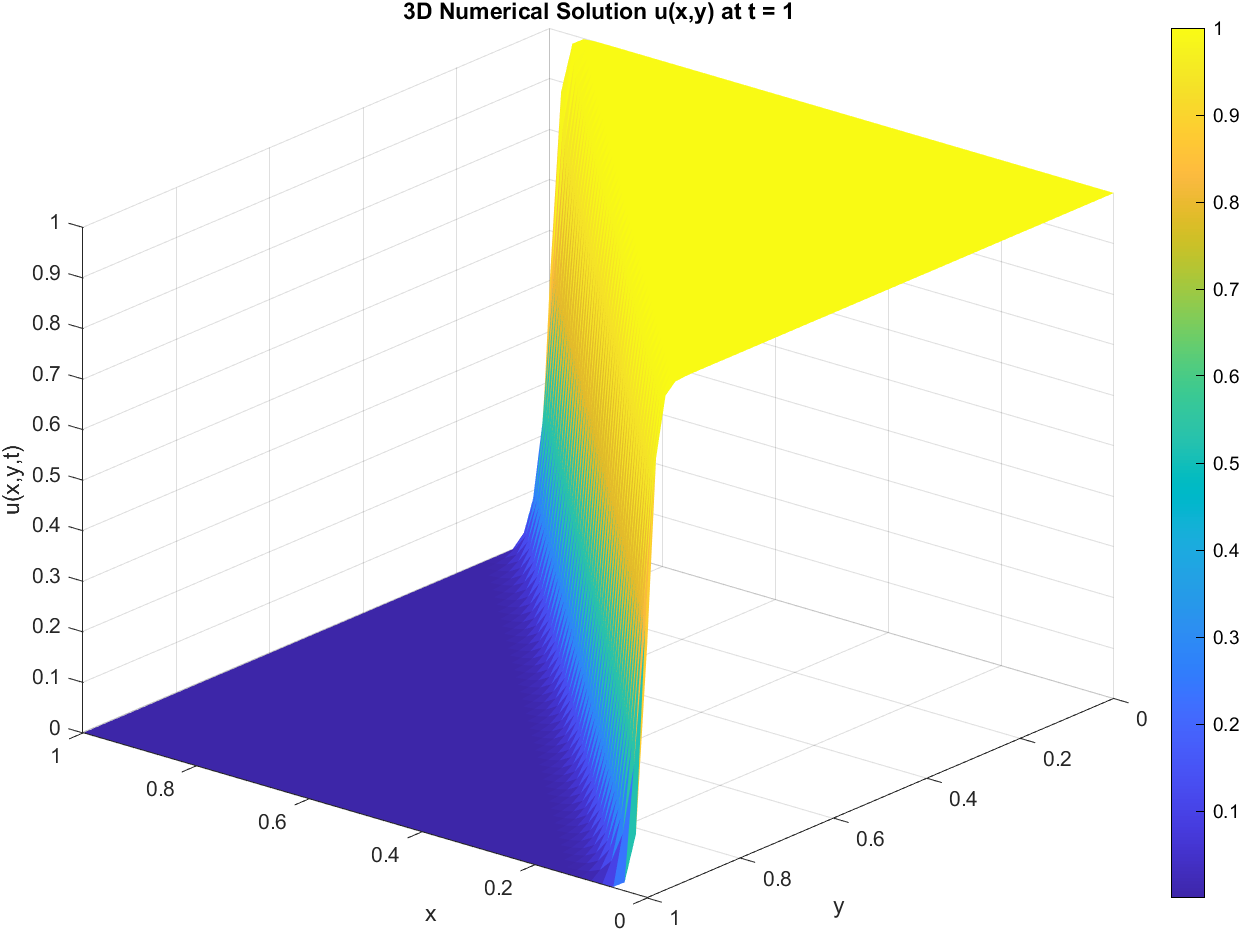}
			\end{minipage}
			\caption{2D Burgers' equation solution at different time levels $t=0$ and $t=1$}
			\label{fig:burgers_2x2}
		\end{figure}
		Figure~\ref{fig:burgers_2x2} illustrates the 2D Burgers' equation solutions at two distinct time levels, \( t=0 \) and \( t=1 \). The initial state (\( t=0 \)) shows the starting profile, while the solution at \( t=1 \) demonstrates the evolution and nonlinear effects, including the formation of sharper gradients and wave interactions over time. These plots highlight the dynamic behavior and dissipative nature of the model.
		
	\end{example}
	\begin{example}
		In the last problem, we consider 2D coupled Burgers' model as
		\begin{align*}
			\frac{1}{\mathrm{R}} \left( \nabla^2 u \right) &= \frac{\partial u}{\partial t} + u \frac{\partial u}{\partial x} + v \frac{\partial u}{\partial y}, \\
			\frac{1}{\mathrm{R}} \left( \nabla^2 v \right) &= \frac{\partial v}{\partial t} + u \frac{\partial v}{\partial x} + v \frac{\partial v}{\partial y}. 
		\end{align*}

		with ID and BDs are taken from the given manufactured solution over  \([0,1] \times [0,1]\):
		\[
		u(x,y,t) = 0.75 - \frac{0.25}{1 + e^{\frac{R(4y - 4x - t)}{32}}}, \quad 
		v(x,y,t) = 0.75 + \frac{0.25}{1 + e^{\frac{R(4y - 4x - t)}{32}}} \quad 
		\]
	\begin{figure}[h!]
		\centering
		\begin{minipage}{0.48\textwidth}
			\centering
			\includegraphics[width=\textwidth]{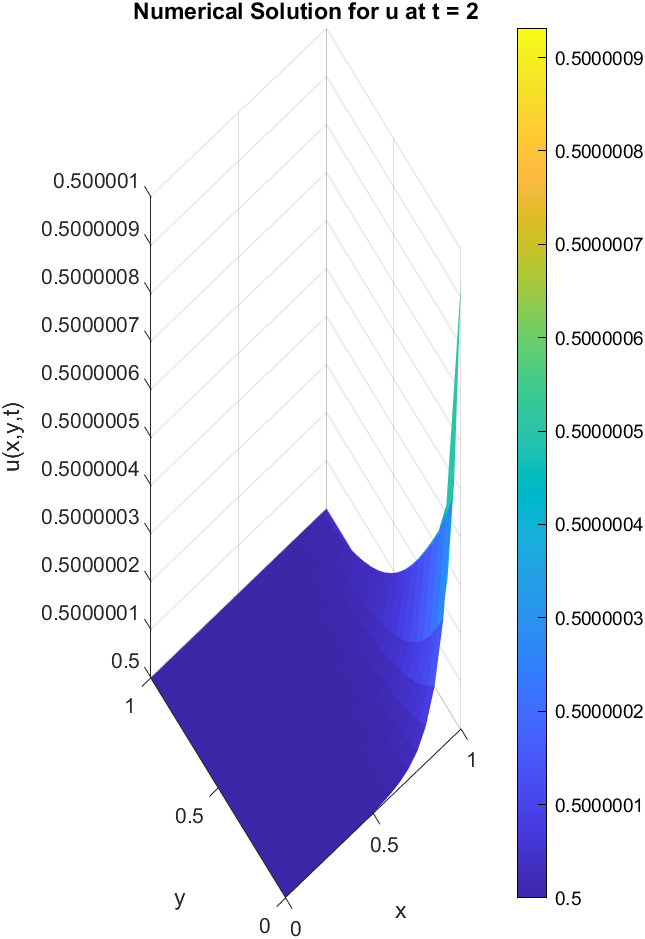}
		\end{minipage}
		\hfill
		\begin{minipage}{0.48\textwidth}
			\centering
			\includegraphics[width=\textwidth]{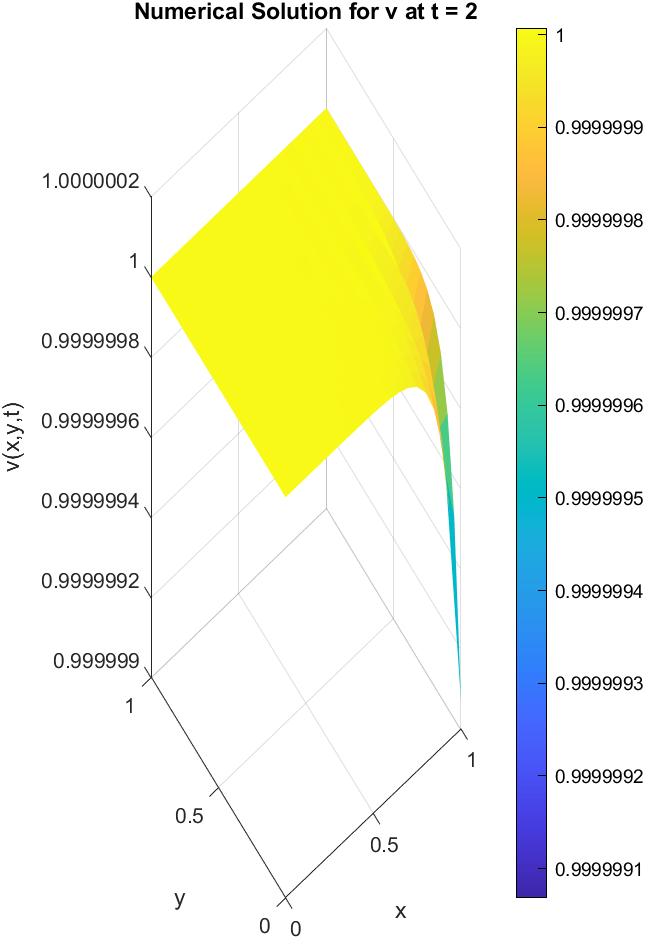}
		\end{minipage}
		\caption{2D Coupled Burgers' equation solution $u(x,y)$ and $v(x,y)$ at time $t=2$}
		\label{fig:burgers_2x2}
	\end{figure}
	\begin{figure}[h!]
		\centering
		\begin{minipage}{0.48\textwidth}
			\centering
			\includegraphics[width=\textwidth]{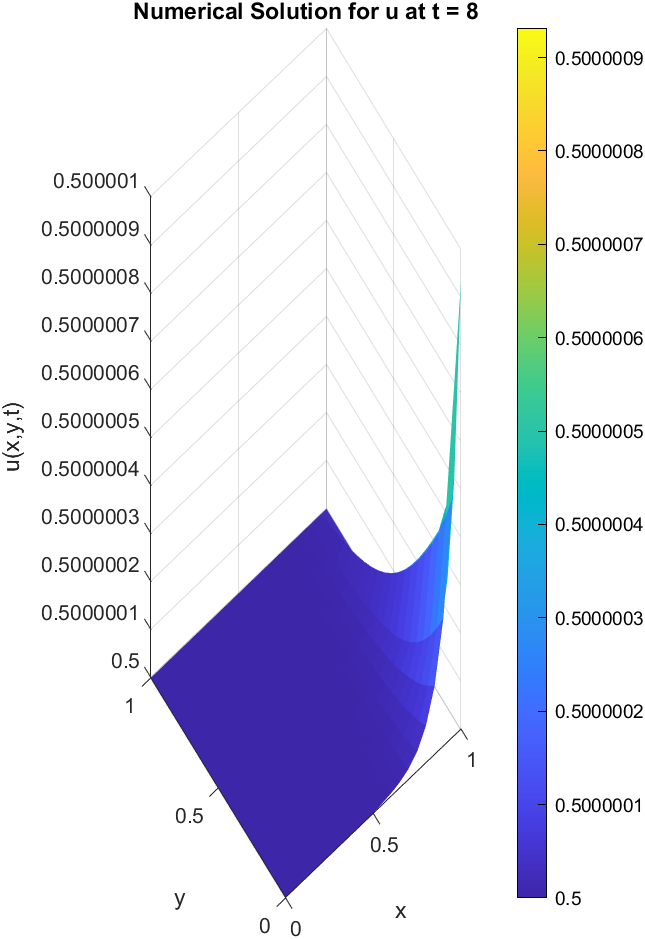}
		\end{minipage}
		\hfill
		\begin{minipage}{0.48\textwidth}
			\centering
			\includegraphics[width=\textwidth]{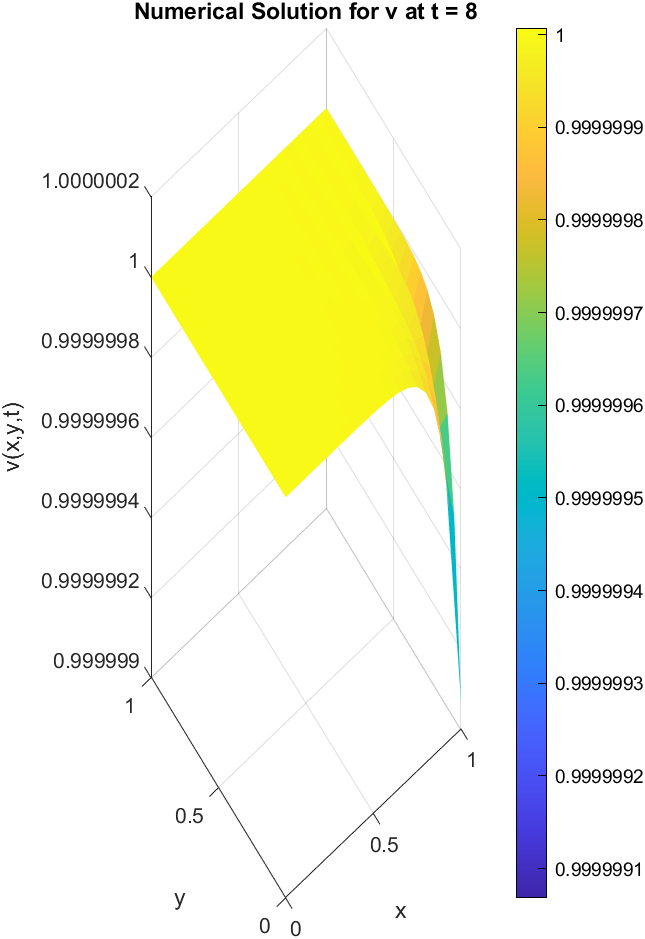}
		\end{minipage}
		\caption{2D Coupled Burgers' equation solution $u(x,y)$ and $v(x,y)$ at time $t=8$}
		\label{fig:burgers_2x2}
	\end{figure}
	Figures~\ref{fig:burgers_2x2} display the solutions of the 2D coupled models for variables \( u(x,y) \) and \( v(x,y) \) at different times. The first figure shows the solution profiles at \( t=2 \), where nonlinear interactions begin to form complex structures. The second set at \( t=8 \) highlights further evolution, with sharper gradients and more pronounced wave patterns, demonstrating the dynamic coupling and dissipative effects over time.
	
\end{example}
\clearpage
\section{Concluding Remarks and Discussion} \label{sec:conclusion}
In this study, PINNs based algorithm have been utilized to simulated the 1D coupled, 2D single and 2D coupled Burgers' models. By embedding the governing PDEs into the loss function, PINNs ensure the learned solution adheres to the underlying physics, even with limited data. The authors conclude the following remark:\\
(i) The mesh-free, differentiable framework eliminates the need for spatial discretization, making it suitable for higher-dimensional systems.\\
(ii) The algorithm demonstrates strong agreement with analytical solutions, as confirmed by the $L^\infty$ and $L^2$ error norms.
Also the algorithm worked for high Reynolds Number $10^6$. \\
(iii) PINN offers better and more effective results than the refs \cite{khater2008chebyshev,rashid2009fourier,bak2019semi,ahmad2019numerical} .\\
(iv) The performance benefits from a flexible network architecture and careful hyperparameter tuning.\\
Overall, this work underscores the effectiveness of PINNs as a scalable and accurate tool for modeling nonlinear PDEs systems, with potential applications in fluid dynamics, reaction-diffusion models, and multi-physics simulations.\\

\textbf{Data availability:} This paper does not include any data that can be made available.\\
\textbf{Declarations}\\
\textbf{Conflict of interest:} The authors declare no conflicts of interest.
\bibliographystyle{abbrv}
\bibliography{myb}
\end{document}